\documentclass{article}

\usepackage{amsfonts}
\usepackage{amsmath}
\usepackage{comment}

\usepackage{microtype}
\usepackage{graphicx}
\usepackage{subfigure}
\usepackage{booktabs} 

\usepackage[numbers]{natbib}
\usepackage{authblk}

\usepackage{hyperref}

\usepackage[accepted]{mlsys2025}

\mlsystitlerunning{MoEBlaze: Breaking the Memory Wall for Efficient MoE Training on Modern GPUs}

\begin{document}

\vskip 0.2in

\twocolumn[
\mlsystitle{MoEBlaze: Breaking the Memory Wall for Efficient MoE Training on Modern GPUs}
\mlsyssetsymbol{equal}{*}

\begin{mlsysauthorlist}
\mlsysauthor{Jiyuan Zhang}{meta}
\mlsysauthor{Yining Liu}{meta}
\mlsysauthor{Siqi Yan}{meta}
\mlsysauthor{Lisen Deng}{meta}
\mlsysauthor{Jennifer Cao}{meta}
\mlsysauthor{Shuqi Yang}{meta}
\mlsysauthor{Min Ni}{meta}
\mlsysauthor{Bi Xue}{tml}
\mlsysauthor{Shen Li}{meta}
\end{mlsysauthorlist}

\vskip 0.05in
\centerline{$~^1 \textit{Meta Platforms Inc},  ~^2 \textit{Thinking Machines Lab}$}

\mlsysaffiliation{meta}{Meta}
\mlsysaffiliation{meta}{Thinking Machine Labs}

\mlsyscorrespondingauthor{Jiyuan Zhang}{jiyuanz@meta.com}
\mlsyscorrespondingauthor{Shen Li}{shenli@meta.com}

\vskip 0.15in

\begin{abstract}

The pervasive “memory wall” bottleneck is significantly amplified in modern large-scale Mixture-of-Experts (MoE) architectures. 
MoE's inherent architectural sparsity leads to sparse arithmetic compute and also introduces substantial activation memory overheads—driven by large token routing buffers and the need to materialize and buffer intermediate tensors.
This memory pressure limits the maximum batch size and sequence length that can fit on GPUs, and also results in excessive data movements that hinders performance and efficient model scaling. We present MoEBlaze, a memory-efficient MoE training framework that addresses these issues through a co-designed system approach: (i) an end-to-end token dispatch and MoE training method with optimized data structures to eliminate intermediate buffers and activation materializing, and (ii) co-designed kernels with smart activation checkpoint to mitigate memory footprint while simultaneously achieving better performance. We demonstrate that MoEBlaze can achieve over $4\times$ speedups and over $50\%$ memory savings compared to existing MoE frameworks. 

\end{abstract}

]

\vskip -0.1cm
\section{Introduction}
\label{intro}


Over the past several decades, processor throughput has advanced much faster than memory bandwidth and latency, creating a persistent “memory wall” that widens the gap between compute and data movement~\citep{memorywall}. In practice, this disparity means that even with ample arithmetic units, end-to-end throughput is often limited by how quickly parameters and activations can be read, written, and exchanged~\cite{arithmetic}.

Mixture-of-Experts (MoE) architectures have reshaped large-scale deep learning by enabling trillion-parameter models at manageable training cost through sparse activation~\citep{shazeer2017outrageously,kaplan2020scaling,hoffmann2022training}. However, the very sparsity that delivers these gains simultaneously lowers compute density because only a subset of experts is active per token. This architectural sparsity, when combined with the scale of distributed training in Large Language Models (LLM), significantly exacerbates memory pressure in modern systems. As models exceed single-device High-Bandwidth Memory (HBM) capacity, training must be distributed across more GPUs and nodes, increasing pressure on device memory bandwidth and interconnect throughput. With the continuous growth in sequence lengths and batch sizes, performance rapidly becomes bounded by the system’s memory and communication subsystems rather than raw FLOPs. In light of this, directly reducing the memory footprint and improving effective bandwidth utilization end-to-end has become critical to break the memory wall for MoE training and achieve efficient model scalings.

While parameter storage often gets the spotlight, activation memory is an equally significant driver of the memory wall during training. In state-of-the-art LLM training at trillion-token scale~\citep{brown2020language, touvron2023llama, gemma2024}, the combination of longer sequences, larger batches, and more complex routing mechanisms leads to a dramatic expansion of the memory buffers required to compact, reorder, and stage intermediate tensors. Consequently, these activation buffers consume a significant portion of GPU memory footprint and bandwidth, directly limiting the maximum batch size and sequence length a system can handle, and thereby capping the model's scalability and training efficiency.

To address this system bottleneck, earlier methods relied on heuristics like token dropping or padding to cap and manage activation buffers~\cite{Samuel59, fedus2021switch}, which, however, often came at the cost of model stability. 
More recent systems are focused on optimizing computation and communication complexity with regards to sparse expert computations~\citep{gale2023megablocks, aminabadi2024turbomoe}. Nevertheless, the auxiliary activation buffers needed for token dispatch and the requirement to pad or materialize intermediate results still contribute a major portion of the overall model memory footprint. 

To address these limitations, we present MoEBlaze, a memory-efficient MoE training framework that drastically improve MoE training memory efficiency without comprising accuracy, while simultaneously achieving better training throughput. 
Concretely, we target two principal sources of activation memory bottleneck: (i) token routing, where conventional implementations allocate large auxiliary per-expert buffers to compact and store activations; and (ii) intermediate activation storage amplified by modern nonlinearities such as SiLU and SwiGLU \citep{glu2020noam,Searching2017Quoc,sigmoid2017stefan}. MoEBlaze is designed to effectively break through the memory wall and maximize the utility of modern GPU architectures to better throughput. Our contributions are:


\begin{itemize}

\item We introduce an efficient end-to-end token dispatch and training method that significantly reduces the intermediate activation buffers for token routing and activation materializing. Our approach avoids both padding and token dropping, reducing memory usage and data movement without sacrificing accuracy while simultaneously achieving better compute efficiency.

\item We introduce efficient data structure and algorithm for above memory-efficient compute scheme that can efficiently leverage GPU's massive parallelism and high bandwidth and avoids complex multi-kernel pipelines.

\item We co-design training kernels with smart activation checkpoint schemes, which can further mitigate the substantial memory footprint associated with modern complex activation functions while achieving better compute efficiency on GPU.

\item Overall our method can achieve over $4$\texttimes{} speedups and over $50\%$ compared memory savings to other state-of-the-art MoE training frameworks across various MoE benchmarks.
\end{itemize}

\section{Background and Motivations}
\label{sec:background}
In this section, we provide a review of the background of MoE training, along with a identification of the key system bottlenecks that currently limit its performance. 

We first define the notations that will be used throughout the rest of the paper. We mainly focus on the token-choice MoE as it has been adopted extensively in production. The MoE computation begins with the token input, which is represented as a vector $\mathbf{x} \in \mathbb{R}^{L \times d}$, where we denote by $L$ the number of routed token instances in a step (e.g., batch size \texttimes{} sequence length), $K$ the number of selected experts per token, $E$ the number of experts, $d$ the model dimension.

\begin{figure*}[htbp]
    \centering
    \includegraphics[width=0.95\textwidth]{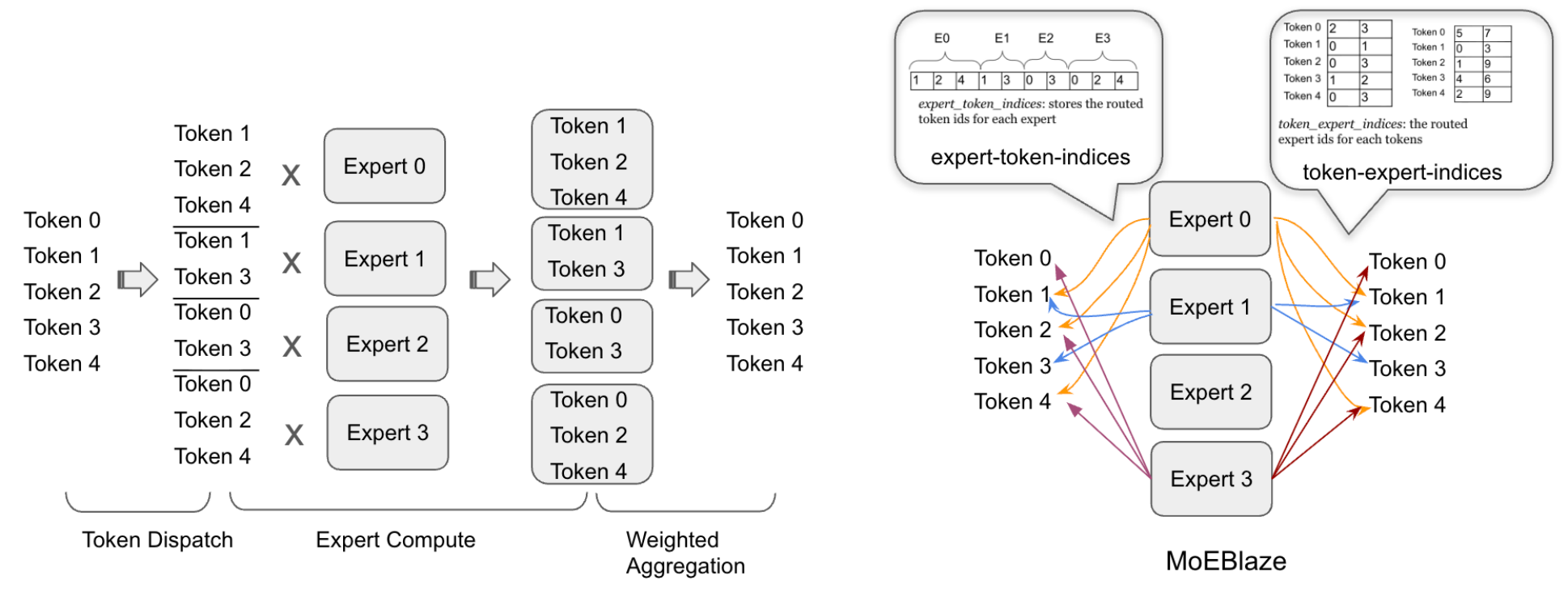}
    \caption{MoE in conventional approach vs. MoEBlaze. Left illustrates the conventional MoE computation, comprising token dispatch, expert computation, and weighted aggregation (details in section~\ref{sec:background}). Right presents the proposed MoEBlaze framework, which introduces memory-efficient token-routing and expert computation (details in section~\ref{sec:memory_efficient_token_dispatch}).}
    \label{fig:moe}
\end{figure*}

\subsection{Gating Network and Token Routing}
\label{sec:gating}
The gating network determines the routing of each input token to the most relevant experts. The network is typically a linear transformation mapping the input dimension $d$ to the number of experts $E$, thereby generating a score for each expert per token. This is followed by a Top-$K$ selection, where expert-ids corresponding to the highest gating scores are collected for each token. The gating output for input $\mathbf{x}$ is defined as:
\[
\text{topk\_experts} = \text{TopK}(\text{softmax}(W_g \mathbf{x}))
\]
where $W_g \in \mathbb{R}^{E \times d}$ are the gating network parameters and $K$ is the number of selected experts per token. The result \textit{topk\_experts} is the list of selected expert-ids by each token.

Following Top-$k$ selection, tokens must be physically routed to their corresponding expert's execution buffer. In conventional implementations, this routing process requires substantial auxiliary memory and extra processing to compact and store the dispatched tokens, which constitutes a critical memory bottleneck.

Earlier work such as Switch Transformers and GShard~\cite{fedus2021switch, lepikhin2021gshard} adopts capacity-limited routing (token-dropping) mechanism to manage token dispatch. Tokens are sorted by their gate score and packed into expert $e$'s buffer; any tokens exceeding $C$ are either dropped or routed to a residual path. A typical choice for capacity is:
\[
C \approx \gamma \cdot \frac{B k}{E}.
\]

where $\gamma$ is the user-defined capacity factor. Capacity-limited routing is amenable to system implements due to fixed-size buffers but comes at the cost of reduced model quality.

More recent literature focuses on dropless routing mechanisms~\cite{rajbhandari2022deepspeedmoe, he2021fastmoe}, which generally yields better model quality. This method ensures every token is processed by an expert, allowing for better model quality and eliminating the need for capacity factor tuning. However, since the number of tokens assigned to each expert is variable, the underlying system must efficiently manage dynamic compute and memory needs. Consequently, recent literature primarily focuses on optimizing the computation with these dynamic and varying-length workloads~\citep{gale2023megablocks, aminabadi2024turbomoe}. 



Nevertheless, a fundamental challenge persists across both token-dropping and dropless routing schemes: current implementations require storing the indices and compacted token data, resulting in memory footprint proportional to $L \times K \times d$. In modern LLM training with longer sequence lengths and higher batch sizes, this leads to a dramatic expansion of the memory buffers.

\textbf{Example:} To illustrate this token dispatch associated activation footprint, we use the example of a real-world MoE model (e.g DeepSeek) for a quantitative study here. For a typical MoE layer in DeepSeek model, it has $L \approx 2$ million tokens, active experts $K = 4$, model dimension $d = 6144$, and using a 2 bytes per element (bfloat16) for the routed token buffer, the memory footprint is:
\[
\text{Mem}_{routing} = L \times d \times k \times 2 bytes \approx 94\text{GB}
\]
We can see a single MoE layer can consume almost a hundred gigabytes of memory for one routing buffer alone. 

\subsection{Expert Feed-Forward Networks (FFNs)}
Following the token dispatch is the Feed-Forward Networks (FFN) computation across experts. Each FFN computation is typically realized as a two-layer Multi-Layer Perceptron (MLP). The first layer projects the input from dimension $d$ to a higher-dimensional hidden space $h$, and the second layer projects back to the output dimension $d'$ (assuming $d'=d$). The total memory required for the parameters across all $E$ experts is $\mathcal{O}(E \times (h \times d + d \times h))$, but the conditional computation paradigm ensures that only $k$ experts are active for each token, maintaining a low computational cost per forward pass. The FFN computation within expert $E_i$ for a given input $\mathbf{x}$ is defined as:
\[
E_i(\mathbf{x}) = W_{2,i} \cdot \sigma(W_{1,i}\mathbf{x})
\]
where $W_{1,i} \in \mathbb{R}^{h \times d}$ and $W_{2,i} \in \mathbb{R}^{d \times h}$, and $\sigma$ is the non-linear activation function (e.g., ReLU, GELU, SwiGLU).

The second principal memory bottleneck stems from intermediate activation storage during the FFN computation. For an active expert, the first linear transformation $W_{1,i}\mathbf{x}$ generates an intermediate activation of size $L_{i} \times h$, where $L_i$ is the number of tokens routed to expert $i$. While the number of active experts is small, the aggregated memory for these activations across all $E$ experts is $\mathcal{O}(L \times h)$ during the forward pass and can be much higher during backpropagation due to the need to store intermediate values for gradient calculation. The choice of activation function (e.g., SwiGLU) will further exacerbate the memory pressure.

\textbf{Example:} We use deepseek's configuration as an example to illustrate the significance of the activation footprint created by FFN computation. We have $L \approx 2$ million, FFN hidden dimension $d = 24576$, and using 2 bytes per element (bfloat16), we can get the activation footprint for the intermediate :
\[
\text{Mem}_{act} = 2L \times h \approx 98 \text{GB}
\]

\subsection{Output Aggregation}
The final stage is output aggregation, where the outputs of the selected experts are combined using a weighted summation to produce the final output for each token. The weights are derived from the gating network's scores. The MoE output $\mathbf{y}$ for an input token $\mathbf{x}$ is:
\[
\mathbf{y} = \sum_{i=1}^{E} g_i(\mathbf{x}) \cdot E_i(\mathbf{x})
\]
where $g_i(\mathbf{x})$ is the gating score for expert $i$, and only the top-$k$ experts have non-zero scores. The memory required for the aggregated outputs is $\mathcal{O}(L \times d)$. Computationally, this involves $\mathcal{O}(L \times k \times d)$ operations, which is generally efficient given the sparsity ($k \ll E$).

\section{Memory-Efficient Token Routing and Training Algorithm}
\label{sec:memory_efficient_token_dispatch}

As detailed in Section \ref{sec:gating}, in token-choice MoE, a gating network assigns each input token to one or more experts. To facilitate efficient token indexing and organization during expert computation, conventional systems compact these tokens into per-expert buffers. This compaction step, followed by the execution of per-expert Multi-Layer Perceptron (MLP) blocks, creates intermediate results kept at the compacted token length before being summed and reduced to the original token length at the output. Crucially, this separation and intermediate storage introduces significant activation buffers throughout the entire MoE training. In this section, we present our memory-efficient routing and expert computation algorithm that substantially reduces the auxiliary activation footprint while also allowing for efficient MoE training.

Given the input as an activation tensor of shape $(L, d)$, the core idea of our algorithm is to leverage auxiliary index lists, generated during the token dispatch step, to track routing decisions and perform on-the-fly token accessing and result reduction throughout the Mixture-of-Experts (MoE) computation. Concretely, our fused kernel operates as follows: 1. it consumes the gating decisions and builds the expert-token index lists and other associated indexing structures, 2. it performs the expert MLP computations using on-the-fly gathers from the original, unpermuted activation tensor, guided by the expert-token index list, and 3. The expert summation then uses the token-expert index list to directly sum and reduce the MLP results into the final output tensor. By directly accessing the input and storing only the final result, we eliminate many intermediate activations that are typically required for materialized token routing in other papers. The token-expert index list, which only stores token and expert IDs, is extremely lightweight. Moreover, this approach allows us to tightly fuse the token/expert indexing with computation, opening possibilities for overlapped memory access and computations. This is particularly advantageous on modern hardware like the latest H100 GPUs, achieving better resource utilization and faster speed.  

Below, we detail the forward and backward passes of our proposed method. Data structures and the methods for efficiently building them will be explained in next section.

\subsection{Forward Pass}
\paragraph{Token Dispatch:} In the token dispatch step, we do not create dedicated buffer for routed tokens. Instead we generate several lightweight indexing data structures based on the gating scores produced in the preceding gating stage. These structures include: the per-expert token list, which tracks the token-IDs assigned to each expert; and the per-token expert list, which stores the expert-IDs chosen for each token. No memory is allocated or preserved for materialized routed token activations at this stage.

\paragraph{Expert Computation:} We perform the expert computation MLPs with on-the-fly gathers from the original unpermuted activation tensor utilizing the indices recorded in the per-expert token list. To maximize memory efficiency, only the intermediate result between the two back-to-back MLPs (i.e., the output of the first MLP) is buffered for the backward pass.

\paragraph{Output Aggregation:} The final results from the experts are  aggregated to produce the final \((L, d)\) output. As we do not store the activation buffer for the materialized token dispatch result, this summation is tightly fused with the 2nd MLP computation and we directly leverage the per-token expert list to perform on-the-fly reduction into output tensor.

\subsection{Backward Pass}
The backward pass takes the gradient of the \((L, d)\) tokens and propagates it back through the inverse of the forward steps. The conventional backward process for expert summation relies on the routed token activation buffer to perform an "expansion" or materialization of the \((L, d)\) gradients to the \((L \times k, d)\) "routed gradient tokens" before backpropagation through the MLP experts. However, our proposed approach avoids this intermediate expansion step by using the same reverse mapping indices. 

\begin{enumerate}
    \item \textbf{Expert Summation Backward:} Using the token-mapping structure derived from the dispatch metadata, the \((L, d)\) gradient tensor is mapped back to the \((L \times k, d)\) routed gradient tokens. This is done via an efficient operation that `scatters` the output gradient to the corresponding locations in the materialized intermediate MLP result tensor.

    \item \textbf{Expert Computation Backward:} Next, the gradients flow backward through the MLPs. The previously checkpointed intermediate result between the two back-to-back MLPs will be used here when computing the weight gradients.

    \item \textbf{Token Gradient Accumulation:} Finally, the gradients with respect to the input tokens are accumulated from all experts. This step sums the contributions from the \(k\) experts each token was routed to, producing the final \((L, d)\) gradient tensor for the input activations. As we do not have the activation storage the materialized routed token result, we also leverage the token index data structure to perform on-the-fly reductions.

\end{enumerate}

\section{Efficient and Parallelizable Dispatch and Data Structures}

\subsection{Data Structures}

We define the key data structures needed for the memory efficient MoE training algorithm we mentioned above. 


\begin{itemize}
 \item  \textbf{$\mathrm{expert\_token\_indices}$}: A compact tensor storing the indices of tokens assigned to each expert, concatenated across all experts. In the token-choice MoE training, each token chooses $k$ experts, thus the \textit{expert\_token\_indices} has size $L \times k$.  This list is fundamental for the experts to retrieve their designated input tokens.
 \item $\mathrm{expert\_token\_offsets}$: An array of length $E + 1$ storing the exclusive prefix sums of token counts per expert. For expert $i$, the indices of its assigned tokens reside from $\mathrm{expert\_token\_offsets}[i]$ up to $\mathrm{expert\_token\_offsets}[i{+}1]-1$.
\item $\mathrm{token\_expert\_indices}$: $\mathrm{token\_expert\_indices}$ is basically the inverse mapping of $\mathrm{expert\_token\_indices}$. It stores the routed expert-ids for each token which are ordered by the token IDs. Its shape is also $L \times k$. This list is need for coalesced indexing into the intermediate materialized results (e.g., between two back-to-back MLPs) when processing tokens per expert.
\item $\mathrm{token\_index\_map}$: A $L\times k$ compact tensor that stores the routed token positions in the $\mathrm{expert\_token\_indices}$ list. It is logically grouped by the original token ID $i \in L$, allowing a token to efficiently find and gather its $k$ expert outputs from the intermediate buffer for the final combination step.
\end{itemize}

\begin{figure}[htbp]
    \centering
    \includegraphics[trim=0.8cm 6cm 0cm 7cm, clip=true, width=0.5\textwidth]{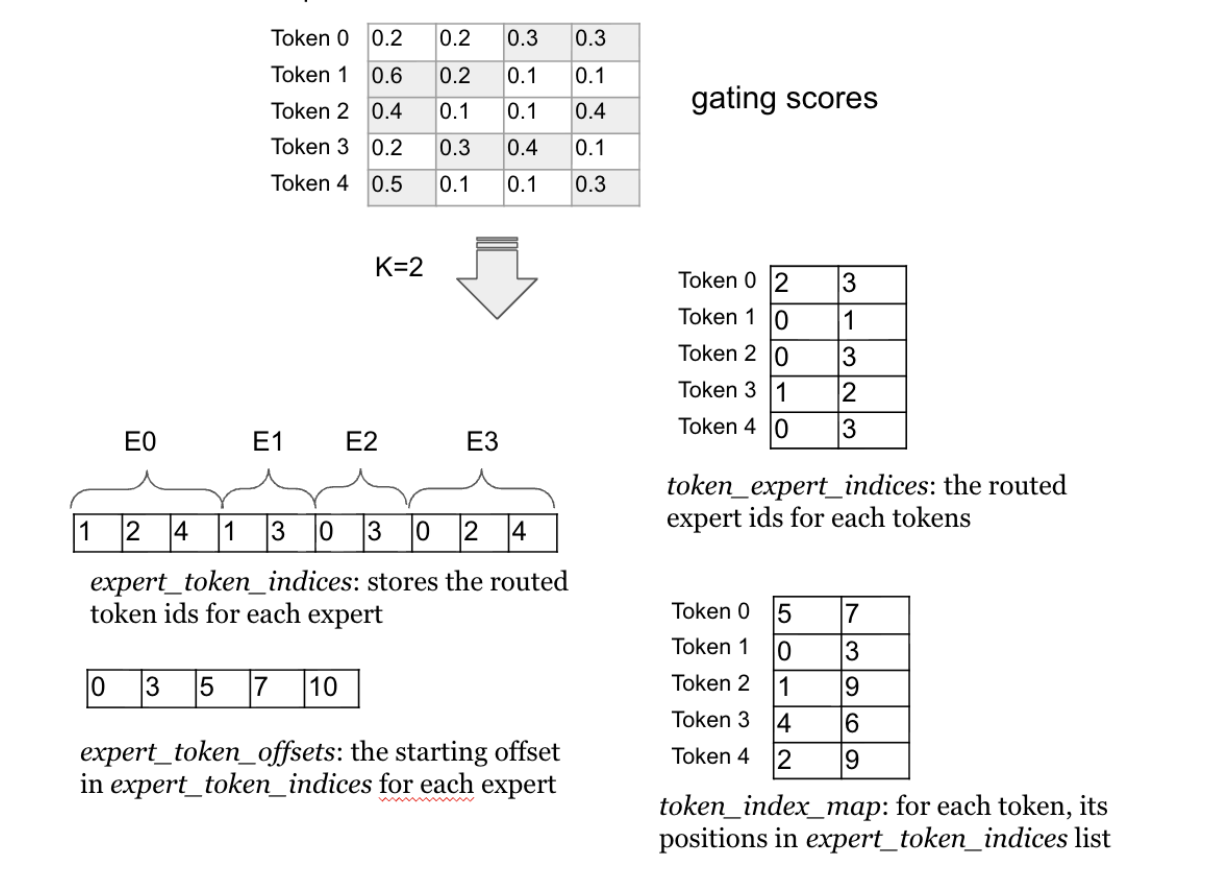}
    \caption{Data structures for the memory-efficient MoE training.}
    \label{fig:indices}
\end{figure}

\textbf{Example.} Figure~\ref{fig:indices} demonstrates the data structure for an example of \(L{=}6\) tokens, \(E{=}4\) experts, and $k=2$ activated experts. From the gating score matrix, we obtain per-token's assignment as: 

Token 0: expert\{2, 3\}; Token 1: expert\{0, 1\}; Token 2: expert\{0, 3\}; Token 3: expert\{1, 2\}; Token 4: expert\{0, 3\}. Concatenate the tokens' assignment together, we will get the $\mathrm{token\_expert\_indices}$ as:
\begin{align}
&\mathrm{token\_expert\_indices} = [2, 3, 0, 1, 1, 2, 0, 3], \nonumber
\end{align}

Similarly we can get for each expert the routed tokens: Expert 0: token\(\{1,2,4\}\); Expert 1: token\(\{1,3\}\); Expert 2: token\(\{0, 3\}\); Expert 3: token\(\{0, 2, 4\}\). Concatenate them together, we get the $\mathrm{expert\_token\_indices}$ and $\mathrm{expert\_token\_offsets}$
\begin{align}
& \mathrm{expert\_token\_indices} = [1, 2, 4, 1, 3, 0, 3, 0, 2, 4], \nonumber \\
& \mathrm{expert\_token\_offsets} = [0, 3, 5, 7, 10] \nonumber
\end{align}
The $\mathbf{\mathrm{token\_index\_map}}$ stores the positions of each token within the concatenated experts' token list.  For example, $\mathrm{token\_index\_map}[0]=\{5, 7\}$ as token\_0 is routed to 2 experts ($k=2$) and placed in the 5th and 7th position of the $\mathrm{expert\_token\_indices}$.

\subsection{Efficient Dispatch Data Structure Construction}

We now detail the methods to efficiently construct the aforementioned data structures. The construction process presents a challenge: the inherent design of the expert-centric data structures requires a many-to-one mapping where multiple tokens are assigned to the same expert. Utilizing a naive approach would result in severe thread-level write contention on the GPU architecture, thereby compromising performance.




One solution is to rely on a sorting-based approach for building the token dispatch. This method flattens all tokens' top-$k$ choice results (topk\_experts) into a 1D array of length $Lk$ containing (expert\_id, token\_id) tuples. The array is then globally sorted by expert\_id to group tokens, followed by index recovery to reconstruct token order and compute per-expert ranges. 

This sorting procedure, while conceptually simple, introduces severe performance bottlenecks at scale. Sorting is implemented as multi-pass radix sort on GPUs, which requires several global-memory passes proportional to key width, forcing frequent global-memory passes and moving $O(Lk)$ data multiple times. This results in a actual high complexity and poor resource utilization on GPUs. Furthermore, this global ordering step limits fine-grained parallelism, forces a multi-kernel dispatch pipeline (multi-pass sorts, segmented scans, index recoveries etc) with high kernel launch latencies. These limitations motivate a more efficient, GPU-friendly approach.

To this end, we introduce an efficient method that replaces complex global sortings and organizations with parallelizable builds upon local index construction that map well to GPUs.
The method is a simple 3-step process with each step designed to be atomic-free and parallelized on GPU which can minimize expensive global-memory passes and avoids complex multi-kernel pipelines. Below we will go over the details of the three steps.

\paragraph{Build Dense Token-Expert Map} 

In the first step, we construct a dense bitmap denoted as $\mathrm{dense\_token\_map}$ to encode the top-$k$ token-to-expert routing. For each token \(i\), we consider its top-\(k\) assigned experts \(\{e_{i,0}, \dots, e_{i,k-1}\}\). For each gate slot, we set \(\mathrm{dense\_token\_map}[i, e_{i,k}]\) to $i$. All other entries remain unset.

The construction of the encoding map is highly parallelizable on the GPU. We initiate the process by allocating an $\mathbf{L \times E}$ dense map and launching the kernel over the CTA grid. The parallelism is managed by assigning each warp a disjoint tile of token rows ($i$) from which it loads the $\mathbf{top\_k}$ expert results. Each \((i,e)\) pair is written out at most once because expert IDs per token are unique; This guarantees no intra-warp collisions.

\paragraph{Compute Expert Lengths}

Leveraging the constructed $\mathbf{\mathrm{dense\_token\_map}}$, the next step is to efficiently compute the lengths and offsets for the sparse token-ID list for each expert. We launch a custom kernel with the CTA grid mapped across the columns (experts) of $\mathbf{\mathrm{dense\_token\_map}}$. Each CTA is dedicated to a single expert $e_i$ and counts the non-zero entries (token-to-expert assignments) within that column. The use of warp-level reductions aggregates the row-wise sums within the CTA, producing the $\mathbf{\mathrm{expert\_lengths}}$ array. The value $\mathrm{expert\_lengths}[e_i]$ represents the final number of tokens routed to expert $e_i$. Following the length computation, the $\mathbf{\mathrm{expert\_offsets}}$ are derived by applying a prefix sum over the $\mathrm{expert\_lengths}$ array outside the initial counting kernel.

\paragraph{Route Indices to Gates}



This 3rd step involves generating the per-expert token id list $\mathbf{\mathrm{expert\_token\_indices}}$, which serves as the input for subsequent MLP computations. To achieve a compact, per-expert concatenation of indices in a contention-free manner on the GPU, we employ a two-phase process centered around generating a location map. This map specifies the final destination position-ID for every non-zero entry in the $\mathbf{\mathrm{dense\_token\_map}}$ within the $\mathbf{\mathrm{expert\_token\_indices}}$ list. Once the location map is built, a simple parallel kernel reads elements from $\mathbf{\mathrm{dense\_token\_map}}$ and writes them directly to their calculated, corresponding positions in $\mathbf{\mathrm{expert\_token\_indices}}$, guaranteeing full parallelism without atomics.

The construction of the location map can be challenging. We utilize a two-step strategy to ensure its atomic-free construction: (i). tile-level scan: We launch one CTA per expert. Threads within the same CTA process contiguous tokens assigned to that expert in $\mathbf{\mathrm{dense\_token\_map}}$. They first compute the tile-level counts within shared memory, followed by an exclusive scan operation (prefix sum) performed locally inside the CTA. (ii). The resulting CTA-local exclusive scan counts then add with the expert's pre-computed global $\mathbf{\mathrm{expert\_offsets}}$. This addition yields the correct, final position-ID in the concatenated indices array.

\section{Training–Kernel Co-Design for End-to-End Efficiency}
\label{sec:checkpoint}


This section details our approach to jointly optimize the Mixture-of-Experts training kernels and smart activation checkpointing method to address the memory issues associated with some advanced activation methods. 

\subsection{SwiGLU MoE and the Memory Bottleneck}
Modern MoE training has increasingly adopted advanced non-linear activations such as \emph{SiLU} and \emph{SwiGLU} in place of ReLU/GELU. Prior work shows these activations provide smoother nonlinearity, which can improve optimization stability and lead to better empirical accuracy on large-scale language tasks. While numerically favorable, these activations introduce more complex compute and larger memory footprint during training. We take the SwiGLU activation as an example. The SwiGLU activation is defined as:
\[
\text{SwiGLU}(\mathbf{x}; W_1, W_2) \;=\; \text{SiLU}(\mathbf{x}W_1) \;\cdot\; (\mathbf{x}W_2),
\]
where $\text{SiLU}(u) = u \cdot \sigma(u)$ and $\sigma$ is the \textit{sigmoid} operation. For an MoE layer with $E$ experts, each implementing a $\text{SwiGLU}$ Feed-Forward Network (FFN), a routed batch of tokens $x \in \mathbb{R}^{L \times d}$ for a single expert induces two projections: $a = x W_1\in \mathbb{R}^{L\times h}$ and $b = x W_2\in \mathbb{R}^{L\times h}$. This is followed by the element-wise operations $\text{SiLU}(a)$ and the final product $\text{SiLU}(a)\odot b$.

\subsection{Activation Checkpoint and Kernel Codesign }
In conventional kernels, the forward pipeline necessitates materializing multiple intermediates in order to accommodate the backward computation  (e.g., in the SwiGLU example, it includes the two GEMM outputs $a$ and $b$, the sigmoid $\sigma(a)$, $\text{SiLU}(a)$, and the final product). These intermediate results are written to and subsequently read from global memory, which incurs non-negligible overheads. As models and batch sizes scale, this incurs significant memory traffic and storage costs, which quickly becomes a non-negligible bottleneck.

To mitigate the observed memory pressure, we present a joint optimization of the MoE training flow and its underlying GPU kernels that reduces the activation memory footprint and memory traffic without sacrificing performance. 
Our optimization is based on below observations:
\begin{itemize}

\item Computation of activation functions is generally memory bandwidth bound on modern GPUs due to two primary reasons: 1) activation function's computation is mostly point-wise operations and modern GPU is highly capable of such operations, 2) in LLM training, we are usually handling the case where the number of tokens is far larger than the embedding dimension $L \gg d$. Operations on matrices of this tall-and-skinny shape are generally memory bandwidth bound on GPUs.

\item While activation computation is computationally light, its memory footprint is surprisingly significant. This is particularly true for complex, modern activation functions, which requires materialization and saving many intermediate, point-wise results for the backward pass. The resulting memory allocation is substantial, scaling linearly with the batch size, sequence length, and FFN dimensions. In the context of today's trillion-token training environments, the memory required to store these activations can be prohibitive.
\end{itemize}

Based on this observations, we propose the joint activation checkpoint and kernel fusion approach.
Our approach fuses the two first-layer projections in SwiGLU with the activation epilogue, and applies activation checkpoint in the specialized path to “break the memory wall” arising from intermediate activation storage.

To reduce activation traffic and kernel launch overhead, we fuse both first-layer projections and the $\text{SwiGLU}$ epilogue into a single kernel.
The kernel consumes non-materialized routed tokens, loads the input $x$ only once, streams it through both $(W_1, W_2)$ GEMMs simultaneously, computes $\text{SiLU}(a)$ in register/shared memory, and immediately performs the multiplication with $b$, writing only the final output to global memory.

This "epilogue fusion" eliminates global writes of $a, b$ and subsequent re-reads for elementwise operations, effectively moving computation from the memory-bound domain to the compute-bound domain where possible. It also halves the input reads of $x$ compared to separate kernels for each projection.

During backward, fusing the two first-layer projections implies that gradients w.r.t. the shared input \(x\) from both paths must be aggregated. Rather than allocating two separate activation buffers and stitching them, our implementation computes the two branches’ activation derivatives in a fused fashion and aggregates gradients in-place via tiled reductions—completely eliminating temporary global buffers.

On top of the fusion, we further applied the activation checkpoint strategy -- where we skip saving the SwiGLU intermediate result (SiLU) during forward. Instead, we adopt a recomputation strategy during the backward pass, leveraging the fact that the $\text{SiLU}$ function is computationally inexpensive (e.g elementwise operation), and are heavily memory bandwidth bounded on modern GPUs.

\begin{algorithm}[H]
    \caption{Fused $\text{SwiGLU}$ MoE Training}
    \label{alg:fused_swiglu}
    \textbf{Input:} Input Tokens $X \in \mathbb{R}^{L \times d}$, Projection Weights $W_1, W_2 \in \mathbb{R}^{d \times h}, W_3 \in \mathbb{R}^{h \times d}$,  \\
    \textbf{Output:} Output $Y_{\text{out}} \in \mathbb{R}^{L \times d}$, Gradients $\nabla W_1, \nabla W_2, \nabla W_3, \nabla Z$
    \begin{algorithmic}[1]
        \STATE // Forward module for Swiglu MoE training
        \STATE \textbf{Procedure FusedForward($X, W_1, W_2, W_3$)} 
        \STATE // \textit{Load input tokens once}
        \STATE \textbf{Load} $X$ 
        \STATE // \textit{1st MLP projection:}
        \STATE //   \textit{Compute $A$ and $B$;} 
        \STATE //   \textit{$\text{SiLU}(A)$ and $Y_{\text{swi}}$ computed in-kernel}
        \STATE //   \textit{$\text{SiLU}(A)$ is transient}
        \STATE $(A, B), Y_{\text{swi}} \leftarrow \text{Fused\_SwiGLU}(Z, W_1, W_2)$ 

        \STATE $Y_{\text{out}} \leftarrow Y_{\text{swi}} W_3$ 
        \STATE \textbf{Store} $A, B, Y_{\text{swi}}$ 
        \STATE // \textit{Store activations and $\text{SwiGLU}$ output for backward}
        \STATE \textbf{Return} $Y_{\text{out}}$

        \STATE
        \STATE \textbf{Procedure FusedBackward}
        \STATE ($Y_{\text{out}}, \nabla Y_{\text{out}}, W_1, W_2, W_3, A, B, Y_{\text{swi}}$) 

        \STATE // \textit{Gradient for final projection}
        \STATE $\nabla W_3 \leftarrow Y_{\text{swi}}^T \nabla Y_{\text{out}}$ 
        \STATE // \textit{Backpropagate gradient to $\text{SwiGLU}$ output}
        \STATE $\nabla Y_{\text{swi}} \leftarrow \nabla Y_{\text{out}} W_3^T$ 

        \STATE // \textit{Load stored activations}
        \STATE \textbf{Load} $A, B$ 

        \STATE // \textit{Recomputes $\text{SiLU}(A)$ to save memory}
        \STATE $S_{\text{recomp}} \leftarrow \text{SiLU}(A)$ 
        
        \STATE // \textit{Derivative w.r.t $A$}
        \STATE $\nabla A \leftarrow \nabla Y_{\text{swi}} \odot B \odot \nabla \text{SiLU}(A)$ 
        \STATE // \textit{Derivative w.r.t $B$}
        \STATE $\nabla B \leftarrow \nabla Y_{\text{swi}} \odot S_{\text{recomp}}$

        \STATE $\nabla W_1, \nabla W_2 \leftarrow \text{FusedBwdW}(X, \nabla A, \nabla B)$

        
        \STATE $\nabla X \leftarrow \text{FusedBwdX}(\nabla A W_1^T, \nabla B W_2^T)$ 
        \STATE \textbf{Return} $\nabla W_1, \nabla W_2, \nabla W_3, \nabla Z$
    \end{algorithmic}
\end{algorithm}

\subsection{Putting It Together: E2E Training on Swiglu MoE}
Algorithm~\ref{alg:fused_swiglu} summarizes the end-to-end training process for an MoE model utilizing the SwiGLU activation function. The pseudo-code specifically demonstrates the integration of activation checkpoint and kernel fusion detailed in section~\ref{sec:checkpoint}. While the low-level implementation of the fused kernels is omitted for brevity, the high-level methodology is derived from the memory-efficient token dispatch explained in section~\ref{sec:memory_efficient_token_dispatch}.

\section{Experiments}
In this section, we benchmark MoEBlaze against the current state-of-the-art sparse training system, Megablocks~\ref{gale2023megablocks}, demonstrating significant improvements in both training speed and memory efficiency across a range of representative Mixture-of-Experts (MoE) configurations.

\subsection{Experiment Setups}
We conducted all experiments on a single NVIDIA H100 Tensor Core GPU. The software stack utilizes PyTorch $2.0.1$ and CUDA $12.1$. We measure end-to-end training time for a single MoE layer, focusing on the Sparse-to-Sparse computation phase. We evaluate performance with two different activation functions: ReLU (Rectified Linear Unit) and SwiGLU (Swish-Gated Linear Unit).

We selected a set of seven representative MoE configurations that explore varied dimensions for the input hidden size ($d$), the number of experts ($E$), the top-$k$ tokens routed to experts, and common training parameters (sequence length $L$ and batch size $B$). The specific configurations, which mimic common settings in large language models, are detailed in Table 1.

\begin{table}[t]
\caption{MoE configurations used in experiments. The FFN hidden dim is set to be four times the input dimension ($\text{ffn\_hidden\_size} = 4 \times \text{input\_d}$).}
\label{sample-table}
\begin{center}
\begin{small}
\begin{sc}
\begin{tabular}{lccccc}
\toprule
\textbf{}      & Input\_d & Experts \# & K & Batch & Seq\_len \\
\midrule
\text{conf1} & 512   & 4   & 1 & 32 & 2048 \\
\text{conf2} & 1024  & 8   & 2 & 32 & 2048 \\
\text{conf3} & 1024  & 16  & 4 & 32 & 2048 \\
\text{conf4} & 2048  & 16  & 4 & 32 & 1024 \\
\text{conf5} & 512   & 16  & 4 & 32 & 1024 \\
\text{conf6} & 1024  & 16  & 4 & 16 & 1024 \\
\text{conf7} & 2048  & 8   & 4 & 16 & 512  \\
\bottomrule
\end{tabular}
\end{sc}
\end{small}
\end{center}
\vskip -0.1in
\end{table}

\subsection{Baselines and Metrics}
Our primary baseline for comparison is Megablocks, a system that optimizes MoE training through custom kernels and efficient token dispatch, serving as the industry standard for high-performance sparse training.

We evaluate performance using two key metrics: (1) Training Speed: measured as the speedup factor of MoEBlaze relative to Megablocks in an end-to-end single training pass. The training time includes both forward and backward passes, but we exclude optimizer updates as optimizer is irrelavant to both approach designs. Higher values indicate better performance. (2) Activation Memory Consumption: measured as the total memory allocated to save the intermediate activation tensors for given inputs. To measure the activation memory, we utilize PyTorch's saved tensor hooks to trace and calculate the exact activation space allocated during model training with the given input configuration.

\subsection{Memory Efficiency in MoE Training using SiLU}
\begin{figure}[htbp]
    \centering
    \includegraphics[width=0.5\textwidth]{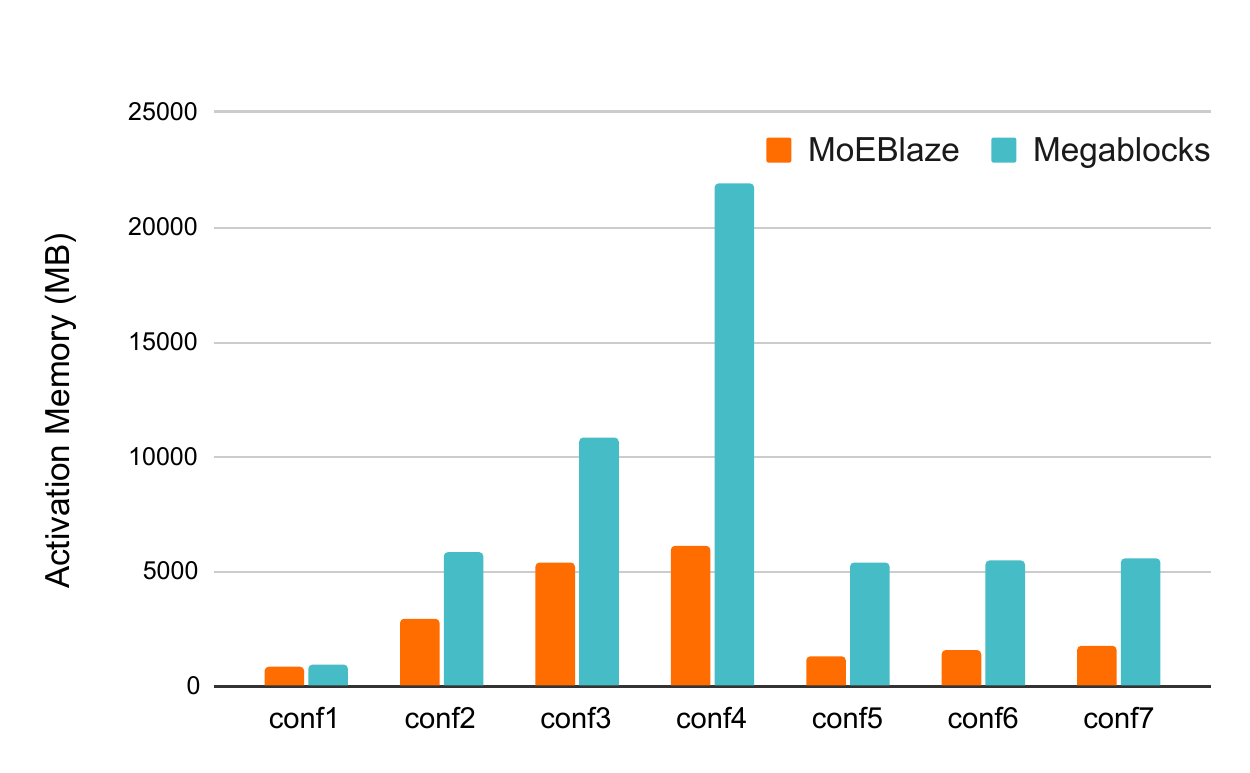}
    \caption{Activation memory footprint comparison between MoEBlaze and Megablocks across the set of MoE configs defined in Table~\ref{sample-table} using SiLU activation function.}
    \label{fig:mem}
\end{figure}

As shown in Figure~\ref{fig:mem},  MoEBlaze consistently and significantly reduces activation memory consumption compared to the Megablocks baseline across all tested configurations.

The memory savings achieved by MoEBlaze are especially significant in configurations characterized by large input dimensions and high expert counts, such as $\text{conf4}$. Specifically, MoEBlaze requires only $6,100 \text{ MB}$ of memory, achieving a nearly $3.6\times$ reduction compared to the $22,000 \text{ MB}$ consumed by Megablocks. For smaller configurations (e.g., $\text{conf1}$), the activation memory saving is less pronounced, which is expected since the savings scale proportionally with the sequence length $L$ and the number of activated experts $k$, both of which are small in $\text{conf1}$ ($k=1$). This substantial reduction in peak activation memory is a direct outcome of two core system innovations: (1) a more memory-efficient token dispatch mechanism that minimizes intermediate buffer allocations, and (2) the adoption of smart recomputation within our custom activation checkpoint scheme. 

\subsection{Training Speed in SiLU-based MoEs}
\begin{figure}[htbp]
    \centering
    \includegraphics[width=0.5\textwidth]{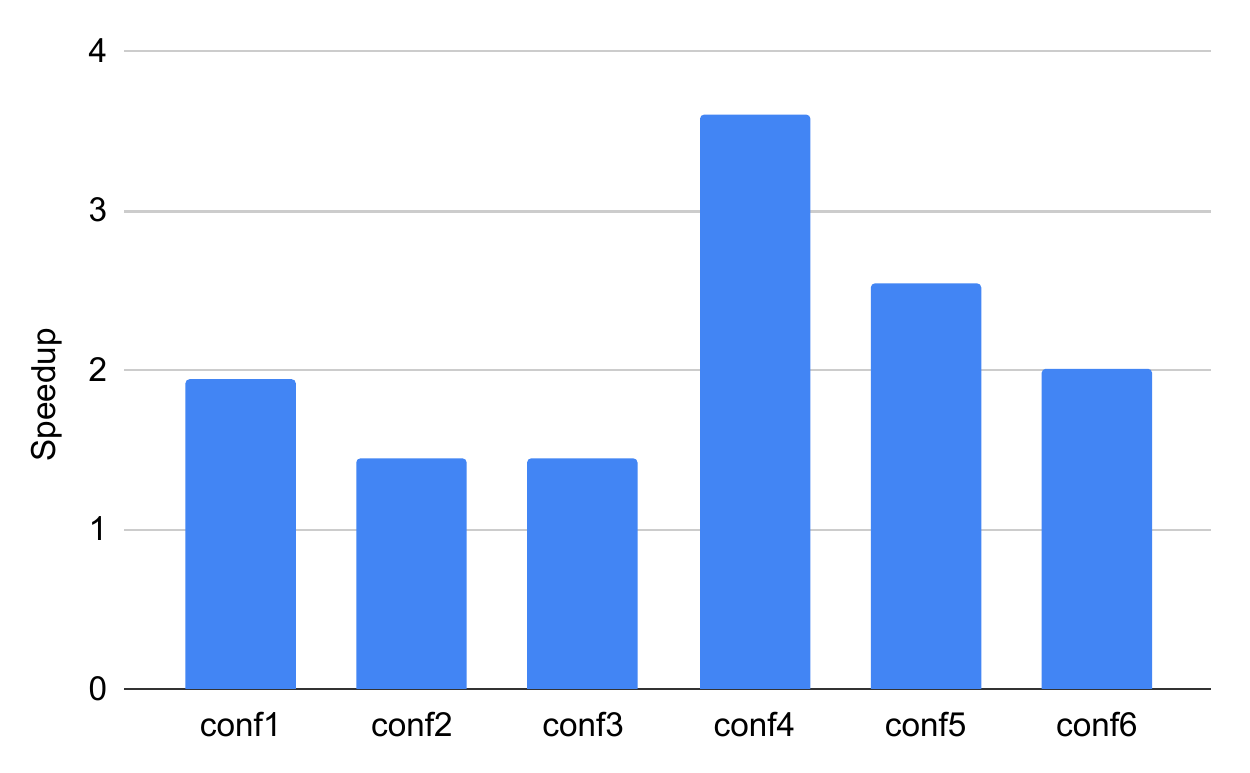}
    \caption{Speedups of MoEBlaze w.r.t to Megablocks on the set of configurations in Table~\ref{sample-table} using SiLU as the activation function.}
    \label{fig:time}
\end{figure}

Figure~\ref{fig:time} illustrates the training speedup of MoEBlaze over Megablocks over the given configurations.  MoEBlaze achieves notable performance gains, showing a speedup factor of $1.4\times$ to $3.7\times$.

The maximum speedup is achieved at $\text{conf4}$ ($D_{input}=2048$, $E=16$, $L=1024$, $B=32$), demonstrating that MoEBlaze scales particularly well with larger model dimensions. This training speeds are attributed to three factors: (1) our highly optimized token dispatch implementation, which reduces the latency overheads associated with expensive token dispatch and permute operations; (2) the efficient data dispatch construction kernels, which is very light-weight and runs rapidly on GPUs, avoiding the expensive multiple-passes kernel in other sorting-based approaches and greatly eliminating the CPU-side bottlenecks. (3) the fused kernel for the batched-GEMM computations that effectively leverages H100's latest hardware acceleration features such as warp-group matrix multiplication, tensor memory accelerator, etc. 

\subsection{Memory Efficiency in MoE Training with SwiGLU}
\begin{figure}[htbp]
    \centering
    \includegraphics[width=0.5\textwidth]{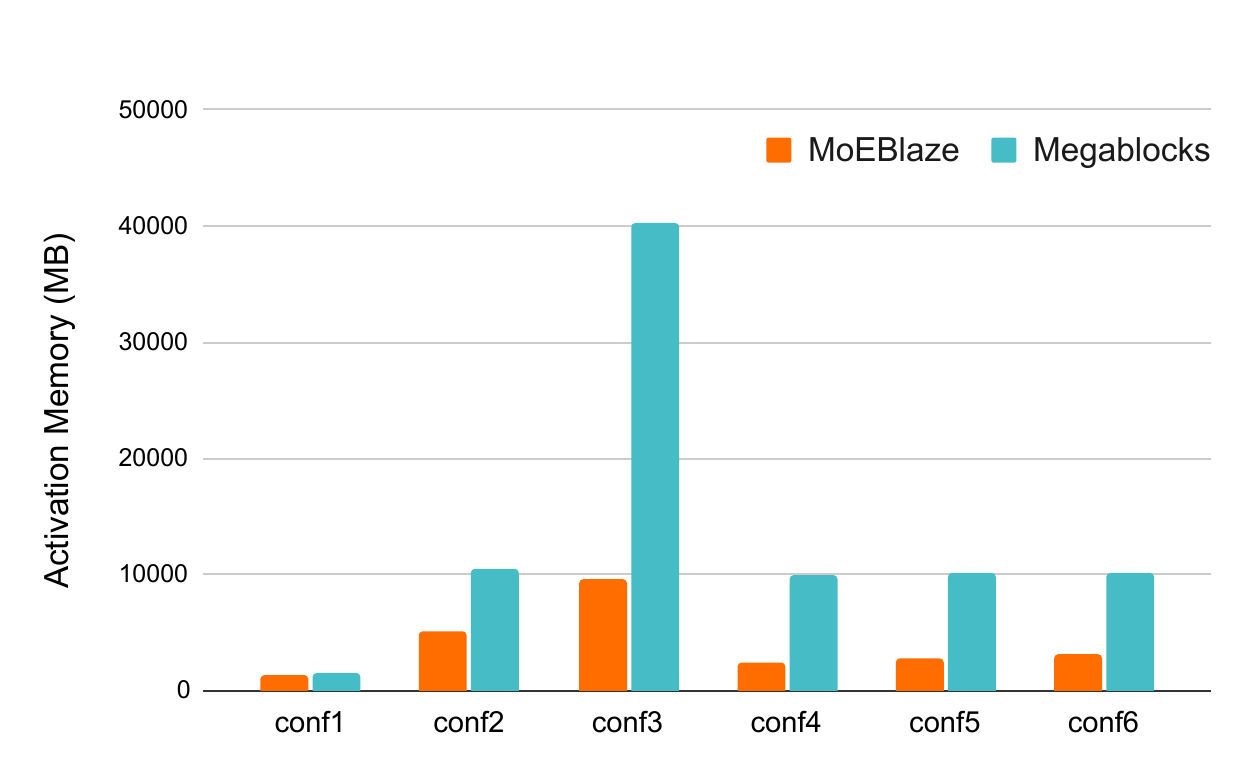}
    \caption{Activation memory footprint comparison between MoEBlaze and Megablocks across the set of MoE configs defined in Table~\ref{sample-table} using SwiGLU activation function.}
    \label{fig:swiglu_mem}
\end{figure}

The SwiGLU activation function inherently requires higher memory usage due to the additional gating and element-wise multiplication operations. Figure~\ref{fig:swiglu_mem} shows the memory consumption comparison under the SwiGLU setting. MoEBlaze maintains a substantial memory advantage over Megablocks, with peak activation memory often less than half of the baseline's usage. For instance, in configuration $\text{conf3}$, Megablocks requires over $40,000 \text{ MB}$, while MoEBlaze is contained to approximately $10,000 \text{ MB}$. This consistent $4\times$ reduction in memory pressure confirms that our memory-efficient dispatch and smart recomputation schemes are highly effective even for more complex activation functions.

\subsection{Training Speed in SwiGLU-based MoEs}
Figure 4 presents the speedup of MoEBlaze relative to Megablocks when using the SwiGLU activation. Compared to the ReLU results, the speedup factors are generally higher and more consistent, ranging from $2\times$ to $6.2\times$. 
The increased relative speed is a result of two factors: (1) The more complex computation in SwiGLU exposes greater opportunities for MoEBlaze's highly fused kernels to outperform the baseline; and (2) the memory-bandwidth savings from our activation optimization are more critical in the SwiGLU case, where intermediate activation sizes are larger and more compound, thereby reducing the excessive global memory accesses through smart kernel fusion and recomputation allows MoEBlaze to execute the whole kernel faster.

\begin{figure}[htbp]
    \centering
    \includegraphics[width=0.5\textwidth]{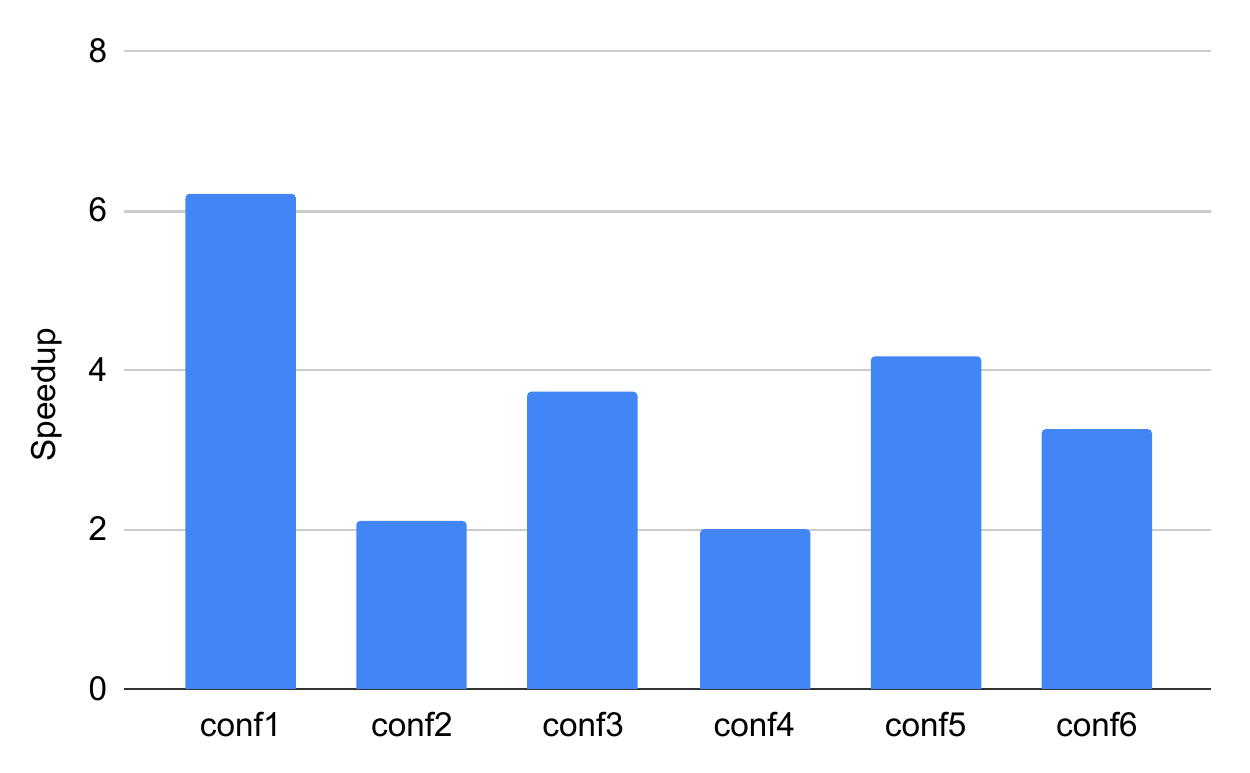}
    \caption{Speedups of MoEBlaze w.r.t to Megablocks on the set of configurations in Table~\ref{sample-table} using SiLU as the activation function.}
    \label{fig:swiglu_time}
\end{figure}

\section{Related Work}

\paragraph{MoE architectures and scaling.}
Scaling laws and empirical studies established that model performance improves predictably with compute, dataset size, and parameter count~\citep{kaplan2020scaling,hoffmann2022training}. GPT-3 demonstrated few-shot generalization at 175B parameters~\citep{brown2020language}, while follow-on models explored scaling in parameters, training data, and context length (e.g., Gopher at 280B~\citep{rae2021gopher}; PaLM at 540B~\citep{chowdhery2023palm}). Open foundation models such as LLaMA~\citep{touvron2023llama} accelerated progress by enabling reproducibility and broad evaluation. More recently, proprietary systems improved multimodal integration and long-context capabilities (e.g., GPT-4~\citep{openai2023gpt4}, Gemini~\citep{gemini2023}), while open releases distilled insights from such systems (Gemma~\citep{gemma2024}). These trends increase pressure on both throughput and memory, particularly under longer contexts and larger hidden dimensions.

Mixture-of-Experts (MoE) was popularized for scaling neural networks via sparse conditional computation, initially in recurrent architectures with the Sparsely-Gated MoE formulation~\citep{shazeer2017outrageously}. Subsequent work demonstrated large-scale Transformer-based MoEs with automatic sharding and conditional computation in production-scale systems (e.g., GShard)~\cite{Samuel59}. Switch Transformers replaced top-$k$ expert selection with top-1 to simplify routing and improve throughput~\citep{fedus2021switch}. GLaM explored large-scale MoE training with expert sparsity and strong efficiency–quality tradeoffs~\citep{du2021glam}. In the open-source ecosystem, Mixtral 8$\times$7B employs top-2 routing with strong performance at 32k context~\citep{mixtral2024}, while DeepSeek-V3 reports a 671B-parameter MoE with 37B active parameters per token and efficient large-scale training~\citep{deepseekv3_2024}.

\paragraph{Systems for MoE training and routing.}
Early GPU-first stacks such as FastMoE offered a PyTorch-based distributed MoE system with practical acceleration and multi-node expert placement~\citep{he2021fastmoe}. Tutel proposed Flex, a design enabling runtime-adaptive parallelism and pipelining to handle routing-induced workload variability, showing large speedups across scales~\citep{hwang2022tutel}. DeepSpeed-MoE introduced both training and inference optimizations to support next-generation MoE at scale~\citep{rajbhandari2022deepspeedmoe}. MegaBlocks reformulated MoE computation as block-sparse operations to avoid padding or token dropping and map well to GPUs~\citep{gale2023megablocks}. TurboMoE argued the gating path is a core bottleneck and introduced fused, metadata-driven kernels and data-layout transformations that reduce sparse-compute overheads and improve large-scale throughput~\citep{aminabadi2024turbomoe}.

\paragraph{Routing policies, load balancing, and capacity.}
MoE quality and efficiency depend on the router. The auxiliary load-balancing loss in Sparsely-Gated MoE mitigates expert imbalance~\citep{shazeer2017outrageously}; Switch Transformers adopted top-1 routing plus capacity constraints to reduce compute and simplify gather/scatter~\citep{fedus2021switch}. GShard explored routing and tensor-sharding policies at massive scale~\citep{lepikhin2021gshard}. Open MoE models such as Mixtral use top-2 routing and capacity factors tuned for stability and throughput~\citep{mixtral2024}. DeepSeek-V3 further reports an auxiliary-loss-free strategy for load balancing while scaling training to very large regimes~\citep{deepseekv3_2024}. Across these designs, routing capacity and token dropping vs.\ padding interact with both throughput and memory pressure, particularly at long contexts.

\paragraph{Kernel and performance optimization.}
GPU performance for MoE hinges on data movement minimization and on-chip residency. MegaBlocks leverages block-sparse kernels to avoid wasteful dense padding~\citep{gale2023megablocks}, and TurboMoE fuses gating, scatter/gather, and expert combination with tailored kernels that avoid expensive sparse MMs~\citep{aminabadi2024turbomoe}. Complementary to sparse mapping and orchestration, architecture-conscious fusion can shorten activation lifetimes (e.g., fusing non-linearities such as SiLU/SwiGLU with expert GEMMs) and reduce read/write traffic. Our work (MoEBlaze) advances this line by eliminating per-expert routed activation buffers via compact metadata and co-fusing routing and expert compute in microarchitecture-optimized kernels for current-generation GPUs.

\section{Conclusion and Future Work}
We present MoEBlaze, a fast and memory efficient system for MoE training on GPU. MoEBlaze eliminating the need for materializing large per-expert activation buffers with fused token dispatch and compute kernel designs. Furthermore, MoEBlaze consolidates the MoE computation and activation pipelines to minimize read/write traffic for better memory bandwidth savings and footprint reduction. Our experimental shows that MoEBlaze provides a highly efficient and scalable solution across a range of configurations with over $4\times$ reduction in peak activation memory consumption and delivers end-to-end training speedups reaching $6.2\times$. 

While this paper primarily focuses on single-device performance, we note that the core mechanisms of MoEBlaze are also applicable to distributed settings. As furture work, we plan to extend MoEBlaze to distributed training frameworks and study the optimizations for multi-node, multi-GPU MoE training.

\section{Acknowledgments}
We gratefully acknowledge Carole-Jean Wu for insightful discussions and consultation. We are also grateful to Hongtao Yu for his expertise and support on the Triton language..


\bibliographystyle{mlsys2025}

\end{document}